# Kinematic analysis of a parallel robot for minimally invasive surgery

Calin Vaida[1], Bogdan Gherman[1], Iosif Birlescu[1], Paul Tucan[1], Alexandru Pusca[1], Gabriela Rus[1], Damien Chablat[1,2], Doina Pisla[1,3*]

[1]CESTER, Technical University of Cluj-Napoca, Cluj-Napoca, Romania
[2] École Centrale Nantes, Nantes Université, CNRS, LS2N, UMR 6004, France
[3] Technical Sciences Academy of Romania, 26 Dacia Blvd, 030167 - Bucharest, Romania
{Calin.Vaida, Bogdan.Gherman, Iosif.Birlescu, Paul.Tucan, Alexandru.Pusca, Gabriela.Rus, Damien.Chablat, Doina.Pisla}@mep.utcluj.ro
[*]Corresponding author

**Abstract.** The paper presents the kinematic modelling for the coupled motion of a 6-DOF surgical parallel robot PARA-SILSROB which guides a mobile platform carrying the surgical instruments, and the actuators of the sub-modules which hold these tools. To increase the surgical procedure safety, a closed form solution for the kinematic model is derived and then, the forward and inverse kinematic models for the mobile orientation platform are obtained. The kinematic models are used in numerical simulations for the reorientation of the endoscopic camera, which imposes an automated compensatory motion from the active instruments' modules.

## 1. Introduction

Robotic assisted minimally invasive surgery (MIS) has continuously evolved since its introduction in the 20[th] century [1]. Various robotic solutions are documented in scientific literature, such as the da Vinci system [1,2], Senhance [1,3], Versius [1,4], PARASURG-5M [5], among many others.

All these systems operate within the surgical field using the Remote Center of Motion (RCM) which is located within the trocar, through which the surgical instruments are inserted into the patient's body. Within Single Incision Laparoscopic Surgery (SILS), a particular approach of MIS, the instruments, usually an endoscopic camera and two instruments featuring active functions, are inserted using a single port. Various mechanisms able to work with a RCM have been proposed, [6-7], some of them with an architecturally constrained RCM, [2], [8]. An important aspect in the design of the medical equipment consists in its capabilities to adapt to the personalized needs of the patient's needs or the procedure's requirements, [9]. Reference is made here to the latest practice guidelines [10], which shows that advanced solutions like robotic systems, intraoperative imaging techniques enable the treatment of borderline non-resectable tumors, improving the survival of patients with locally advanced pancreatic cancers.

A modular robotic system – PARA-SILSROB for MIS (SILS) suitable for pancreatic cancer treatment, in which a 6 DOF parallel robot is used to position a mobile platform on which the endoscope and two active medical instruments are positioned using RCM mechanisms has been previously presented, [10]. The authors have also defined the spherical RCM mechanism, with the sphere center in the RCM [11, 12]. The paper presents the kinematic model of the robotic system which enables the orientation of the mobile platform during the surgical field inspection and at the same time preserves the position of the active instruments using the RCM mechanisms, thus increasing the safety of the surgical procedure.

The paper is structured as follows: Section 2 presents the PARA-SILSROB modular robotic system and defines the different motion types in robotic-assisted surgical task; in Section 3 the general kinematic model is derived; Section 4 presents numerical simulations and Section 5 describes the conclusions.

## 2. PARA-SILSROB and the surgical task description

The PARA-SILSROB robotic system experimental model is presented in fig. 1.

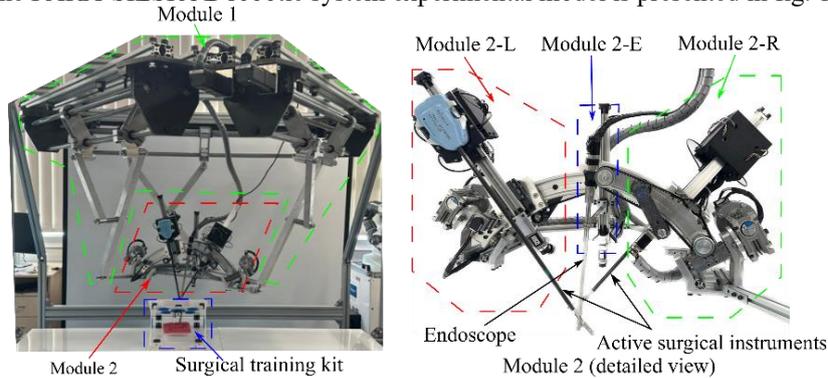

**Fig 1.** The experimental model of PARA-SILSROB.

It consists of two modules:
- **Module 1** – is represented by a 6 DOF parallel robot [11] that guides a triangular mobile platform (module 2), fig. 3. The parallel robot consists of three identical kinematic chains, each having two active prismatic joints ($q_i$, $i=1..6$), three passive revolute joints ($R_{i1}$, $R_{i2}$, $R_{i3}$, $i=1..3$), two cylindrical joints ($C_{i1}$, $C_{i2}$, $i=1..3$) and one spherical joint connecting the mobile platform ($S_i$, $i=1..3$).
- **Module 2** – consists of the mobile platform which holds the active instruments having a 1-DOF insertion/retraction mechanism for the endoscopic camera (module 2-E) and two (symmetrically positioned with respect to the endoscopic camera) 3-DOF insertion and orientation mechanisms holding the left and right active instruments). Module 2-L and module 2-R are based on spherical mechanisms with architecturally constrained remote center of motion (RCM) [13].

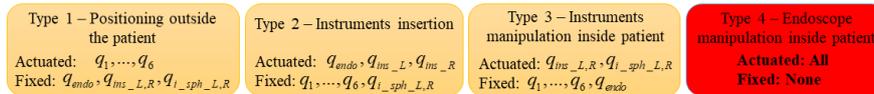

**Fig 2.** The motion types of the PARASILS-ROB robot

The motions performed by the robot in MIS can be grouped in 4 types (Figure 2):
- **Type 1** - the robot moves the mobile platform outside the patient; this motion is used for the relative positioning of the robot with respect to the patient (and the SILS entry ports) where only the robot actuators are active.
- **Type 2** - the surgical instruments are inserted in the surgical field on linear trajectories using the three insertion active joints.
- **Type 3** – the active instruments are manipulated (independently) using the actuators of the 3-DOF modules L/R.
- **Type 4** - the endoscope orientation is changed (by reorienting the mobile platform). As this motion will also reposition the active instruments, they must perform compensatory motions to remain with the tips in the same location (with respect to a fixed coordinate system).

It is important to consider that while the instruments are inserted into the operating field, a rotation about the longitudinal axis of the endoscope produces a rotation of the SILS port at the skin level (which should be avoided). Consequently, following the instruments insertion inside the patient, the rotation of the mobile platform around its Z axis is fixed (orientation parameter $\varphi$=ct. from the kinematic models in [10]). The last type of motion is by far the most complex one and is used to validate the kinematic models presented in the next section.

## 3. The kinematic model of the surgical robot

A general kinematic model for the 6-DOF parallel robot was introduced in [11], characterizing the pose of the point $P(X_P, Y_P, Z_P, \psi_P, \theta_P, \varphi_P)$ as illustrated in Figure 3. Without any loss of generalization, it can be assumed that this point P represents the entry point of the endoscope inside the patient. Based on [11] and the kinematic scheme of the mobile platform, this point P is defined relative to the entry ports used by the spherical mechanisms driving the active instruments (as those points are architecturally fixed). In the *O'X'Y'Z'* system, these coordinates are: $P_{RCM}$ (0,0,0), $RCM_L$ (–10,0,0), $RCM_R$ (10,0,0) imposed by the SILS ports distance.

A general model is derived for the coupled motion of the mobile platform (actuated by the 6-DOF parallel robot - module 1) and the insertion and orientation modules for the surgical instruments. The forward and inverse kinematics are analytically determined from the general model by considering the specific characteristics of the medical task. A kinematic scheme for the parallel robot and a detailed view of the mobile platform with its instruments modules is illustrated in Figure 3, where the points of interest are also pointed out graphically.

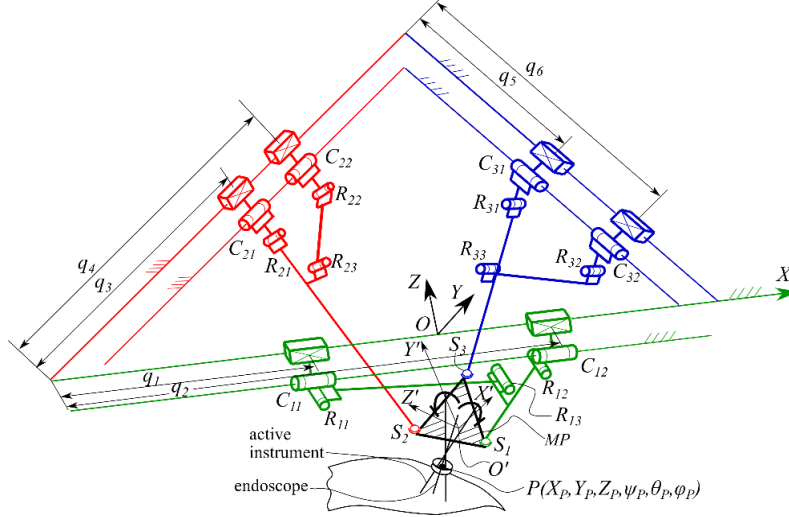

**Fig 3.** A kinematic scheme of the 6-DOF parallel robot

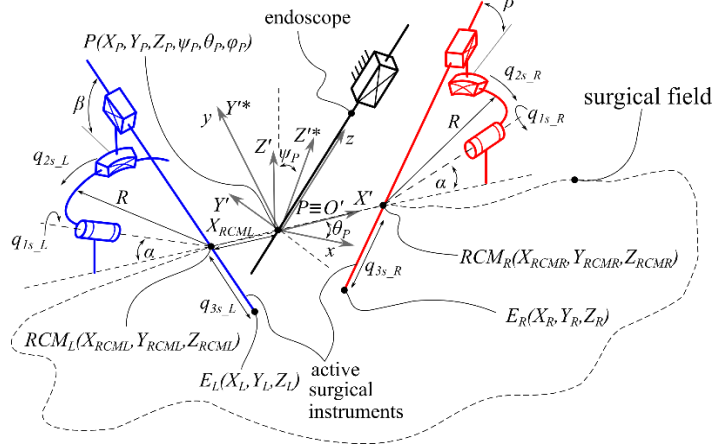

**Fig 4.** A kinematic scheme of the mobile platform (module 2) holding the surgical instruments

To solve the forward and inverse kinematic models, a general transformation matrix is defined to characterize the point P with respect to the 6-DOF parallel robot and the transformation matrix for the $RCM_{L,R}$ in the O'X'Y'Z' mobile system.

$$M_P = \begin{bmatrix} r_{11} & r_{12} & r_{13} & X_P \\ r_{21} & r_{22} & r_{23} & Y_P \\ r_{31} & r_{32} & r_{33} & Z_P \\ 0 & 0 & 0 & 1 \end{bmatrix}, \quad RCM_{L,R} = \begin{bmatrix} 1 & 0 & 0 & X_{RCM} \\ 0 & 1 & 0 & Y_{RCM} \\ 0 & 0 & 1 & Z_{RCM} \\ 0 & 0 & 0 & 1 \end{bmatrix} \quad (1)$$

The $r_{ij}$ terms represent the direction cosines of the X-Y-Z Euler angles which are free of singularity parametrization based on the orientation amplitudes of the mobile platform. To compute the $RCM_L$ and $RCM_R$ coordinates with respect to the $OXYZ$ coordinates system, the next equation is used:

$$RCM_{L\_FIX} = M_{platform} \cdot RCM_L, \text{ and } RCM_{R\_FIX} = M_{platform} \cdot RCM_R \quad (2)$$

Based on the symmetry of the platform from this point on only the left module will be detailed. Using the notations in Figure 4, the transformation matrix for the left spherical module is defined:

$$M_L = R_Y(\alpha) \cdot R_X(q_{s1}) \cdot R_Y(q_{s2}) \cdot R_X(\beta) \cdot T_Z(-q_{s3}) \quad (3)$$

with $R_X$ the rotation matrices around $OX$ axis (by $\alpha$ and $q_{s2}$, respectively), $R_Y$ the rotation matrices around $OY$ axis (by $q_{s2}$ and $\beta$, respectively), and $T_Z$ the translation matrix along the $OZ$ axis (by $-q_{s3}$). To calculate the end-effector coordinates for the active instruments with respect to the fixed coordinate system, a matrix multiplication is used, extracting then the last column:

$$M_{L\_FIX} = RCM_{L\_FIX} \cdot M_L, \text{ and the last vector column:}$$

$$\begin{cases} X_{L\_F} = (X_L + X_{RCM}) \cdot r_{11} + (Y_L + Y_{RCM}) \cdot r_{12} + (Z_L + Z_{RCM}) \cdot r_{13} + X_P \\ Y_{L\_F} = (X_L + X_{RCM}) \cdot r_{21} + (Y_L + Y_{RCM}) \cdot r_{22} + (Z_L + Z_{RCM}) \cdot r_{23} + Y_P \\ Z_{L\_F} = (X_L + X_{RCM}) \cdot r_{31} + (Y_L + Y_{RCM}) \cdot r_{32} + (Z_L + Z_{RCM}) \cdot r_{33} + Z_P \end{cases} \quad (4)$$

In eq. 4, $X_L, Y_L, Z_L$ represent the vector column from $M_L$. Thus, equation (4) defines the general forward kinematic model for the active instruments, as all the terms in the right-hand side are expressions of the inputs.

The inverse kinematic model is computed in a two steps approach:
1. Compute $X_L, Y_L, Z_L$ from eq. 4 (as the fixed coordinates are now known);
2. Solve the vector column of $M_L$ for $q_{1s}, q_{2s},$ and $q_{3s}$.

Before solving the kinematic models for velocities, a short description of the input/output data for the *type 4 motion* is given. When the instruments are inside the patient, the cartesian coordinates of P are fixed, along with the rotation around the Z axis, leaving only two changing variables: $\psi_P$ and $\theta_P$ to reorient the endoscopic camera. As the camera orientation generates a displacement for the active instruments, a compensatory motion must be computed imposing that end-effector coordinates should remain fixed. Thus, to solve the kinematic models for velocities the

6general matrix $A \cdot \dot{X} + B \cdot \dot{Q} = 0$ is used [14,15], where $A$ is the Jacobian matrix computed from eq. (4) with respect to the end-effector coordinates, $X=[\ X_{L\_F},\ Y_{L\_F},\ Z_{L\_F}]^T$, and B is the Jacobian matrix computed with respect to the inputs $Q = [q_{1s}, q_{2s}, q_{3s}, \psi_P, \theta_P]^T$. The expressions for $\dot{q}_{1s}, \dot{q}_{2s}, \dot{q}_{3s}$ can be calculated (in closed form). Similarly, the accelerations are computed from $A \cdot \ddot{X} + \dot{A} \cdot \dot{X} + B \cdot \ddot{Q} + \dot{B} \cdot \dot{Q} = 0$.

## 4. Numerical simulations for the surgical task

During endoscope reorientation achieved using the mobile platform the active instruments are also moving, which can lead to collisions with the internal organs. Numerical simulations have been performed using MATLAB to illustrate how the algorithm generates compensatory motions at the level of the spherical modules which manipulate the active instruments to preserve the current position of their tips (the end-effectors). This motion is specific to the type 4 motion described before. Figure 5 illustrates the time dependent diagrams for the reorientation of the mobile platform (to reorient the endoscopic camera). Figure 6 illustrates the variation of the parallel robot actuators, and Figure 7 illustrates the time dependent diagrams for the spherical module actuators.

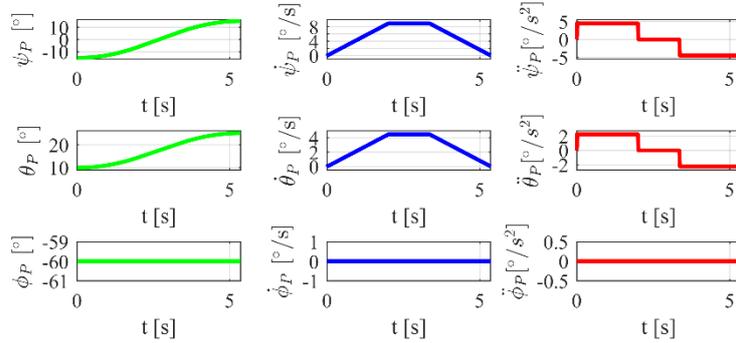

**Fig 5.** Time history diagrams for the mobile platform reorientation

For the simulation, the following parameters were used:
- The initial location of the characteristic point of the mobile platform (the RCM of the endoscope) has the coordinates $P_0$=[15, 20, –500, -15, 10, -60] (mm, °).
- The coordinates of the end-effector are $X_{L\_F}$=50, $Y_{L\_F}$= -50, $Z_{L\_F}$=-620 (mm).
- The endoscope is reoriented with the angles: $\psi = 15°$, $\theta = 25°$. A trapezoidal velocity trajectory has been used, with the maximum angular velocity and acceleration $\omega_{max}$=10°/s and $\varepsilon_{max}$=5°/s², respectively.
- The spherical module must reposition its actuators to compensate for the platform reorientation (preserving the values for $X_{L\_F}, Y_{L\_F}, Z_{L\_F}$).

The following set of geometric parameters, identical with the ones of the experimental model were used for the spherical mechanism: $R = 110$ [mm], $\alpha = 10°$, $\beta$

=10°, while the dimensions of the 6-DOF parallel robot are detailed in [11]. The trajectory preserves the same assembly mode and avoid singular configurations.

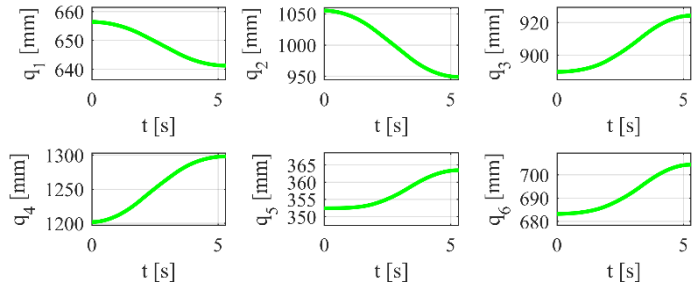

**Fig 6.** Time history diagrams for the 6-DOF parallel robot

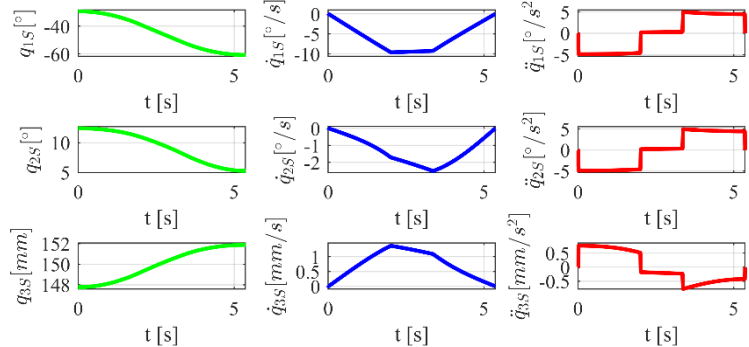

**Fig 7.** Time history diagrams for the left spherical module during mobile platform reorientation.

## 5. Conclusions

An innovative parallel surgical robot for SILS has been presented. The innovative design consists in a modular approach where a 6 DOF parallel robot guides an instrument platform to perform the SILS procedure. The design eliminates the accidental collision of the robotic arms which guides the instruments with the only drawback being the motion of the instruments when the endoscope is reoriented. To solve this, an innovative algorithm is proposed to achieve automatic compensatory motions of the instruments, preserving their fixed location inside the patient when the camera is reoriented. The kinematic model of the robot has been developed, targeting the increase of the surgical procedure safety for cancer treatment. By careful term manipulation the inverse and forward kinematics of a general kinematic model has been achieved, describing the motion of the mobile platform and the equations for the kinematics of the instruments sub-modules in closed form. The numerical simulations have validated the proposed kinematic model, allowing their use within the control system of the developed parallel robot.

……8

## Acknowledgements

…………………
This research was supported by the project New smart and adaptive robotics solutions for personalized minimally invasive surgery in cancer treatment - ATHENA, funded by European Union – NextGenerationEU and Romanian Government, under National Recovery and Resilience Plan for Romania, contract no. 760072/23.05.2023, code CF 116/15.11.2022, through the Romanian Ministry of Research, Innovation and Digitalization, within Component 9, investment I8.